\documentclass[10pt,twocolumn,letterpaper]{article}

\usepackage{iccv}              % To produce the CAMERA-READY version
\definecolor{iccvblue}{rgb}{0.21,0.49,0.74}
\usepackage[pagebackref,breaklinks,colorlinks,allcolors=iccvblue]{hyperref}

\def\paperID{5442} % *** Enter the Paper ID here
\def\confName{ICCV}
\def\confYear{2025}

\title{VideoMerge: Towards Training-free Long Video Generation}

\author{Siyang Zhang\\
University of Central Florida\\
{\tt\small si122915@ucf.edu}
\and
Harry Yang\\
HKUST\\
{\tt\small yangharry@ust.hk}
\and
Ser-Nam Lim\\
University of Central Florida\\
{\tt\small sernam@ucf.edu}
}

\begin{document}

% \maketitle

% \begin{figure*}
% \begin{center}
%     \centering
%     % \includegraphics[width=1.0\textwidth]{figures/compare_a_singer_of_a_music_band.png}
%     % \caption{A singer of a music band.}
%     % \label{fig:qualitative_results_singer}

%     \includegraphics[width=1.0\textwidth]{figures/Comparison_human.png}
%     \caption{Original prompt: \textit{``A woman is dancing in indoor garden.''} Our method is able to preserve consistency in human identity in terms of face, clothes, hair style. }
%     \label{fig:qualitative_results_dancer}

% \end{center}
% \end{figure*}

\twocolumn[{
\renewcommand\twocolumn[1][]{#1}%
\maketitle
\begin{center}
    \vspace{-20pt}
    \includegraphics[
    % width=0.58\textwidth,
    % height=0.5\textwidth
    width=1.0\textwidth,
    height=0.5\textwidth
    ]{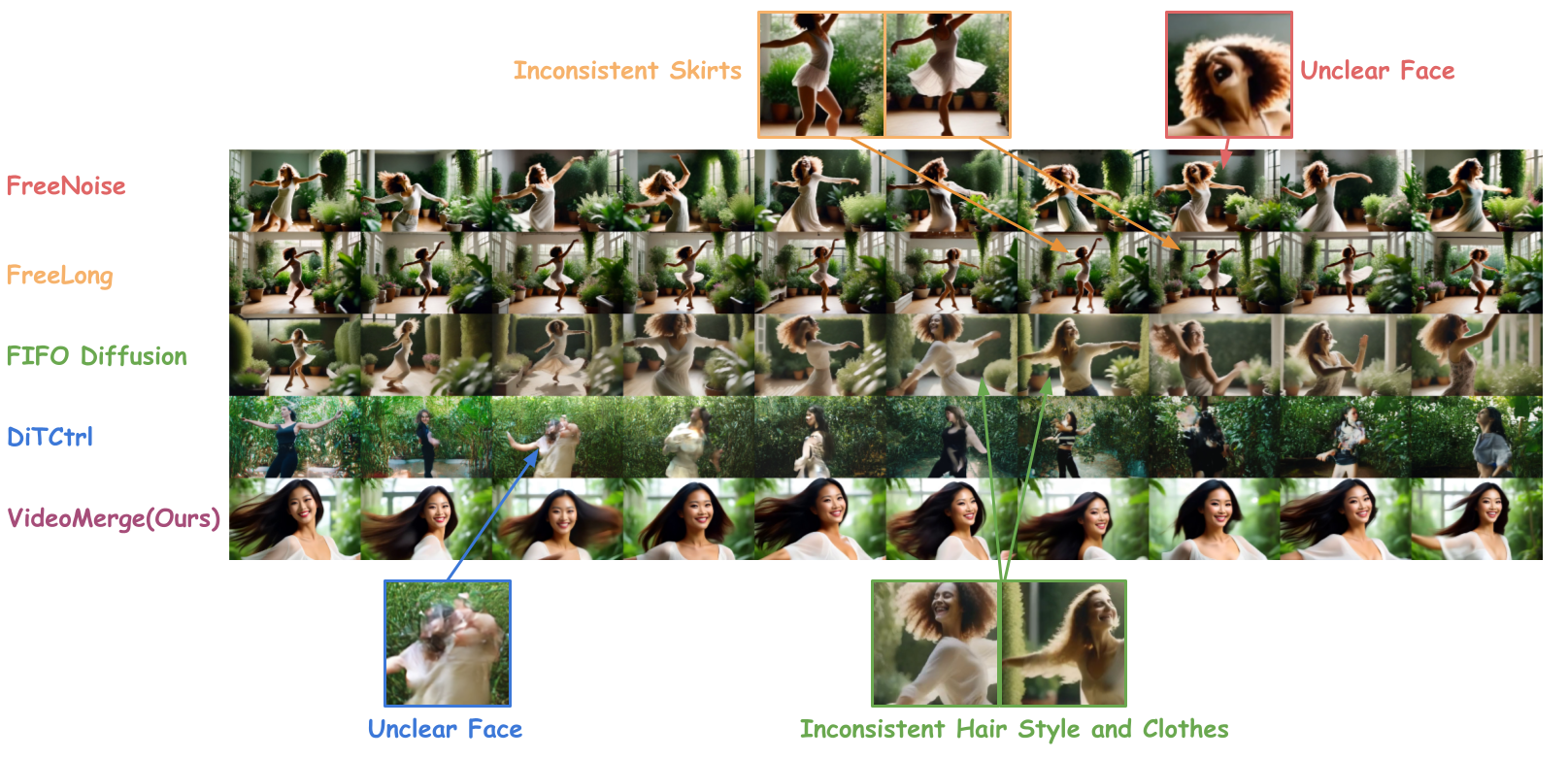}
\vspace{-0.4cm}
\captionof{figure}{Comparison between our proposed \textit{VideoMerge} and other state-of-the-art methods. Original prompt: \textit{``A woman is dancing in indoor garden.''} Our method is able to preserve consistency in human identity in terms of face, clothes, hair style. \textbf{\textit{We provide an extensive list of videos in the supplementary material that clearly demonstrates our superiority over current methods}}.
}
    \vspace{3pt}
    \label{fig:qualitative_results_dancer}
    \end{center}%
}]

\begin{abstract}
Long video generation remains a challenging and compelling topic in computer vision. Diffusion‐based models, among the various approaches to video generation, have achieved state‐of‐the‐art quality with their iterative denoising procedures.
However, the intrinsic complexity of the video domain renders the training of such diffusion models exceedingly expensive in terms of both data curation and computational resources. Moreover, these models typically operate on a fixed noise tensor that represents the video, resulting in predetermined spatial and temporal dimensions (i.e., resolution and length).
Although several high-quality open-source pretrained video diffusion models, jointly trained on images and videos of varying lengths and resolutions, are available, it is generally not recommended to specify a video length at inference that was not included in the training set. Consequently, these models are not readily adaptable to the direct generation of longer videos by merely increasing the specified video length.
In addition to feasibility challenges, long-video generation also encounters quality issues. The domain of long videos is inherently more complex than that of short videos: extended durations introduce greater variability and necessitate long-range temporal consistency, thereby increasing the overall difficulty of the task.
To address these challenges, we propose VideoMerge, a training-free method that can be seamlessly adapted to merge short videos generated by pretrained text-to-video diffusion model. Our approach preserves the model's original expressiveness and consistency while allowing for extended duration and dynamic variation as specified by the user. By leveraging the strengths of pretrained models, our method addresses challenges related to smoothness, consistency, and dynamic content through orthogonal strategies that operate collaboratively to achieve superior quality.

\end{abstract}
    
\section{Introduction}

Recent advancements in video diffusion models have demonstrated remarkable performance in generating high-quality short videos by leveraging extensive training on meticulously curated image and video datasets~\cite{opensora, yuan2024opensoraplan, yang2024cogvideox, kong2024hunyuanvideo}. However, extending these successes to long video generation remains a significant challenge due to two primary issues. First, even under default configurations, the inference cost of video diffusion models is substantial, and naively increasing the noise length for long videos results in extreme memory requirements that exceed typical GPU resource constraints. Second, the use of fixed-size noise tensors during training introduces a train-test discrepancy; forcefully imposing a larger video length can lead to a loss of high-frequency details~\cite{lu2024freelong}. Consequently, there is a consensus among training-free approaches for long video generation that the inference configuration should closely mirror that of training. 

Current training-free methods address long video generation primarily through two strategies: autoregressive generation and latent fusion. Autoregressive methods~\cite{henschel2024streamingt2v, blattmann2023stablevideodiffusion} generate long videos by using previously synthesized results as guidance for subsequent frames. However, this approach tends to accumulate errors by iteratively using synthesized frames, leading to notable quality degradation. In contrast, latent fusion approaches~\cite{qiu2023freenoise, lu2024freelong, cai2024ditctrl, kim2024fifo} avoid relying on synthesized data for guidance and thus circumvent error accumulation. Nevertheless, these methods often struggle to maintain video consistency, particularly in preserving the identities of humans or objects, because they perform denoising on independent noise tensors across different segments.

Despite considerable efforts in prior research~\cite{qiu2023freenoise, lu2024freelong, cai2024ditctrl, kim2024fifo}, the issue of consistency in long video generation remains inadequately addressed. Although existing methods can achieve relatively smooth transitions over extended sequences, they still fail to preserve the global consistency of human or object identities (see Fig.~\ref{fig:qualitative_results_dancer}).

In this work, we introduce \textit{\textbf{VideoMerge}}, a training-free method that adapts a pretrained text-to-(short)video model~\cite{kong2024hunyuanvideo} to generate arbitrarily long videos while maintaining strong global consistency. Our approach addresses the consistency problem through three key innovations. First, rather than randomly initializing a long Gaussian noise which typically exhibits weak global consistency when denoised in short tiles, we reuse a short Gaussian noise of the default size configurated in the pretrained model, extending it to a longer noise tensor. We then blend this extended noise tensor with a new high-frequency component to preserve low-frequency semantic content while introducing dynamic motion details. Second, we propose a novel latent fusion strategy that merges the predicted noise tensors of different tiles, thereby enhancing the smoothness of transitions between denoising segments. Finally, we analyze the influence of text prompts, the sole conditioning input in the generation process, and demonstrate that fine-grained prompt refinement significantly improves the preservation of human or object identities. By integrating these three components, VideoMerge is capable of generating long videos that exhibit both smooth dynamic transitions and strong identity consistency.

% Specifically, our main contributions are as follows: 
% \begin{enumerate} 
%     \item We introduce a new \textbf{latent fusion} method that effectively merges neighboring denoising tiles with smooth transition. 
%     \item We propose a \textbf{long noise initialization} technique that involves shuffling and replicating a short Gaussian noise, which is subsequently augmented with the high-frequency component of a new noise. This approach ensures both consistency and dynamic variability. 
%     \item We analyze the effect of fine-grained prompts and present a \textbf{prompt refining} strategy that further addresses the challenge of preserving human identity in generated videos. 
% \end{enumerate}

Our main contributions are summarized as follows: 
\begin{enumerate} 
    \item We introduce a novel \textbf{latent fusion} method that effectively merges neighboring denoising tiles to achieve smooth transitions. 
    \item We propose a \textbf{long noise initialization} technique that shuffles and replicates a short Gaussian noise, subsequently augmenting it with the high-frequency component of a new noise, thereby ensuring both global consistency and dynamic variability. 
    \item We analyze the impact of fine-grained prompts and present a \textbf{prompt refining} strategy that further addresses the challenge of preserving human identity in generated videos. 
\end{enumerate}

\section{Related Works}

Generative models in computer vision %for vision tasks 
have evolved from GANs~\cite{goodfellow2014generative} 
to diffusion models~\cite{ho2020denoising}.
One of the most widely studied %known application 
problems is Text-to-Image (T2I), popularized by the success of diffusion models in T2I. One of the most effective diffusion models leverages a UNet~\cite{ronneberger2015unet}
structure to model the reverse process of adding Gaussian noise to clear latents. Based on these well developed T2I models' capability of understanding and handling 2D image knowledge, further extension %application such as 
to Text-to-Video (T2V) unleashes the possibility for creative works in the more complex video domain. 

\subsection{Video Diffusion Models}
Early works in video diffusion such as StableVideoDiffusion~\cite{blattmann2023stablevideodiffusion}, VideoCrafter~\cite{chen2024videocrafter2}, and LaVie~\cite{wang2024lavie}) typically extended a pretrained 2D UNet, originally developed for images, by incorporating additional modules to handle temporal attention. These models leverage a 2D+1D (i.e., pseudo-3D) attention mechanism within a 2D latent space derived from a pretrained 2D VAE, thereby obviating the need for temporal compression in the VAE.

Subsequent video diffusion models
(OpenSora~\cite{opensora}, Open-Sora-Plan~\cite{yuan2024opensoraplan}, CogVideoX~\cite{yang2024cogvideox}, Allegro~\cite{allegro2024}, and Hunyuan~\cite{kong2024hunyuanvideo})
% ~\cite{opensora, yuan2024opensoraplan, yang2024cogvideox, allegro2024, kong2024hunyuanvideo}
have predominantly adopted Diffusion Transformers (DiT)~\cite{peebles2023dit} as denoisers. These models utilize full spatio-temporal attention across all dimensions and perform denoising within the latent space of a pretrained 3D-VAE, which more effectively captures features directly from video data. DiT-based diffusion models have established a robust foundation and baseline for text-to-video tasks, achieving state-of-the-art performance.

\subsection{Long Video Generation}
Simply adapting the training methodology~\cite{ho2022video} used for short video generation models to long video models introduces several challenges, including the scarcity of high-quality long video datasets and limited computational resources. Consequently, it is common practice to leverage pretrained short video generation models and employ training-free strategies to adapt them for long video generation.

% Many research works have addressed long video generation using both text-to-video (T2V) and image-to-video (I2V) approaches. In T2V-based methods, where text is the sole conditioning input, previous studies have typically refined the noise and prompt or modified the sampling strategy to mitigate inconsistency issues. Conversely, an intuitive strategy for I2V-based models is to use generated frames as guidance for future generation. However, this approach raises several concerns. First, selecting an appropriate initial guide image is critical; if the chosen image is outside the distribution of the training data, the quality of the generation may be adversely affected. For instance, a model trained on high-quality realistic videos may not generalize well when provided with a Ukiyoe-style image. Second, continuous use of the latest generated frames as guidance can lead to significant error accumulation, with empirical evidence suggesting that even the best I2V models exhibit obvious quality degradation after as few as 10 iterations.
Numerous studies have explored long video generation using text-to-video (T2V)~\cite{chen2024videocrafter2, opensora, yuan2024opensoraplan, yang2024cogvideox, guo2023animatediff} and image-to-video (I2V)~\cite{blattmann2023stablevideodiffusion, yuan2024opensoraplan, yang2024cogvideox, henschel2024streamingt2v} approaches. T2V methods typically refine noise, prompts, or sampling strategies to address inconsistency issues. In contrast, I2V models commonly use generated frames as guidance for future synthesis. However, this strategy presents challenges: selecting an appropriate initial guide image is crucial, as images outside the training distribution can degrade quality, and relying on the latest generated frames for guidance can lead to significant error accumulation, with even top-performing I2V models showing quality degradation after approximately ten iterations.

\subsubsection{Latent Fusion Diffusion with Sliding Windows}
Latent fusion with sliding window~\cite{zhang2024mimicmotion, wang2023genlvideo, bar2024lumiere} is a straightforward technique applied to text-to-video (T2V) generation for long video tasks. It is typically implemented as a linear weighted combination of adjacent video latent tiles (windows). Although this approach can smooth transitions between tiles, it does not fully mitigate substantial motion changes, which can result in object inconsistency and unexpected morphing artifacts. Later works such as FreeNoise~\cite{qiu2023freenoise}
exploit a noise rescheduling idea to maintain temporal consistency. Although such an approach improves overall semantic consistency, there are still artifacts in local details. FreeLong~\cite{lu2024freelong} introduces SpectralBlend Temporal Attention that uses a local temporal attention mask to balance the frequency distribution of long video features, but such computation requires extra memories, which limits the length of the video. %and is thus fails to apply to infinite length. 
DiTCtrl~\cite{cai2024ditctrl}, a recent work built on CogVideoX~\cite{yang2024cogvideox}, introduces a KV-sharing mechanism and a latent blending strategy that facilitate multi-prompt long video generation. However, its primary emphasis on video editing through multiple prompts, which necessitates substantial semantic modifications, often results in a failure to consistently preserve object and human identity.
\subsubsection{First-In-First-Out Diffusion}
First-In-First-Out (FIFO) diffusion~\cite{kim2024fifo} exploits a pretrained UNet-based T2V diffusion model by modifying the denoising procedure. Specifically, it transitions from a simultaneous denoising strategy where all frames are processed using the same noise timestep to a progressive approach in which each frame is denoised with increasing noise timesteps. Although this method maintains memory efficiency comparable to the base T2V inference, it exhibits several limitations. In particular, there is a discrepancy between the fixed time-step schedule used during training and the progressive schedule employed during inference. Moreover, owing to the inherently abstract nature of text prompts and the stochastic characteristics of T2V models, object consistency may deteriorate as the video length increases.
Futhermore, applying FIFO to DiT based diffusion model is problematic. In DiT models, the text embedding is associated with a single timestep, where no explicit fine-grain redistribution of progressive timestep can be applied, resulting in more discrepancy and suboptimal video quality.

\begin{figure}
\begin{center}
    \centering
    \includegraphics[width=0.5\textwidth]{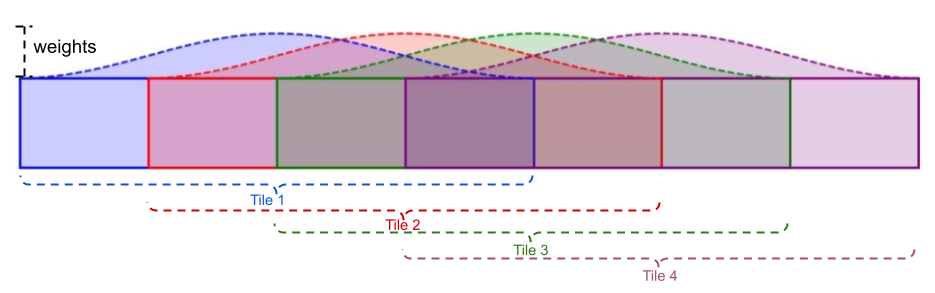}
    \caption{The weight assigned to each frame latent in a latent denoising tile follows a sine curve which allows smooth transition between adjacent tiles.}
    \label{fig:noise_weighting}
\end{center}
\end{figure}

\subsubsection{Autoregressive Model}
Autoregressive generation utilizes previously generated results to predict next result in a sequential manner. There are works such as MagVit-V2~\cite{yu2023magvit2} 
that tokenize videos and allow next video token prediction, enabling long video generation. However, these works are not open-sourced and thus their performances in long video generation remain unclear. 

Open-sourced autoregressive works such as StreamingSVD~\cite{henschel2024streamingt2v} leverage trainable conditional attention module and appearance preservation module to enhance smoothness and consistency, but they still suffer from long range degradation due to error cumulation. 

% \snl{related work is way too long -- almost more than 1 pages}

\begin{figure*}
\begin{center}
    \centering

    % \includegraphics[width=1.0\textwidth]{figures/compare_hummingbird_hawk_moth_flying_near_pink_flowers.png}
    % \caption{Hummingbird hawk moth flying near pink flowers.}
    % \label{fig:qualitative_results_moth}

    \includegraphics[width=1.0\textwidth]{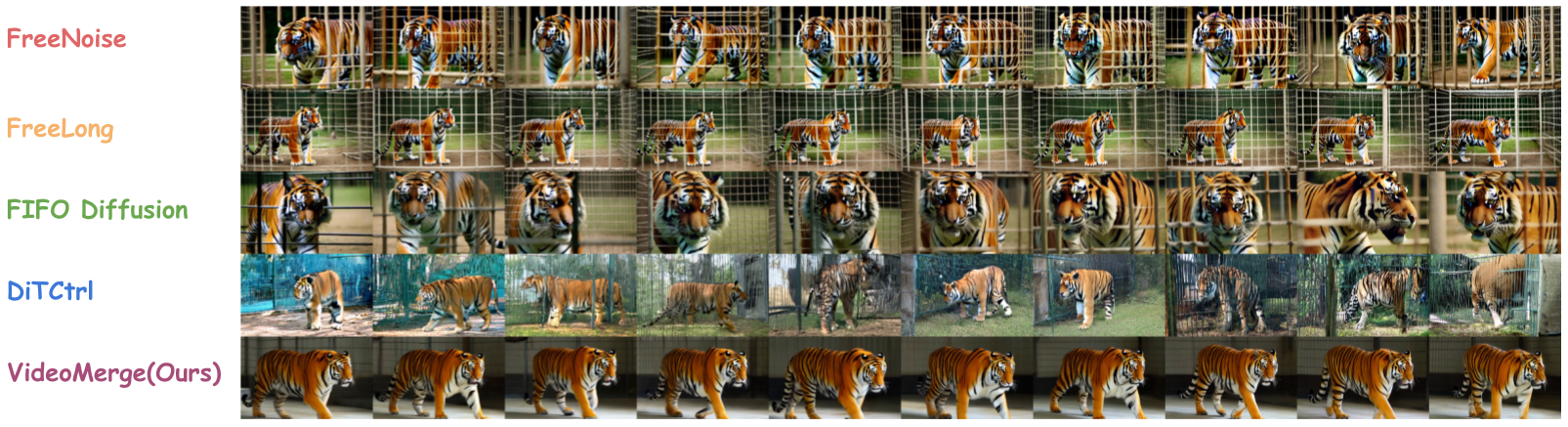}
    \caption{Original prompt: \textit{A tiger is walking inside a cage.}}
    \label{fig:qualitative_results_tiger}

\end{center}
\end{figure*}

\begin{figure*}
\begin{center}
    \centering

    \includegraphics[width=1.0\textwidth]{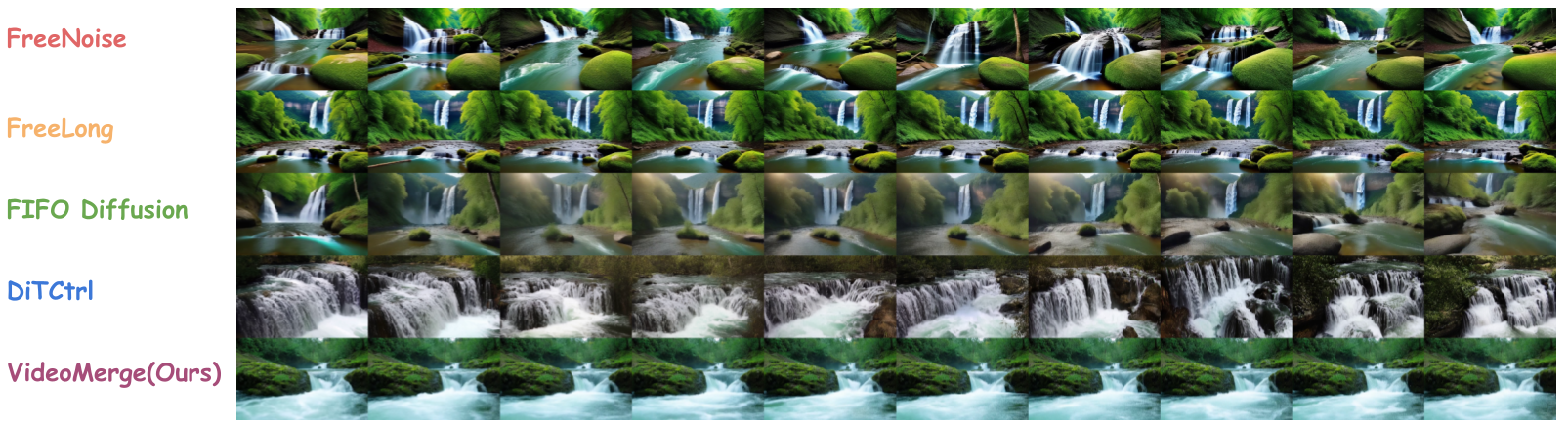}
    \caption{Original prompt: \textit{Beautiful scenery of flowing waterfalls and river.}}
    \label{fig:qualitative_results_waterfall}

    % \includegraphics[width=1.0\textwidth]{figures/compare_a_dark_clouds_over_shadowing_the_full_moon.png}
    % \caption{Dark clouds over shadowing the full moon.}
    % \label{fig:qualitative_results_moon}

\end{center}
\end{figure*}

\section{Methods}
In this section, we present our methods designed to ensure consistency in long video generation. To generate long videos using a pretrained text-to-short-video model while maintaining an acceptable memory budget, it is imperative to align inference parameters closely with the original training configuration, thereby minimizing discrepancies. Also, we need to perform careful condition control so as to keep smoothness, consistency, and identity.

\subsection{Multi-Tile Latent Fusion}
Prior studies~\cite{wang2023genlvideo, bar2024lumiere} have introduced a sliding window strategy to address the inconsistency issue by overlapping consecutive denoising tiles. This approach enhances inter-tile smoothness, as the overlapping regions are computed as a weighted sum of two denoising tiles, thereby inheriting characteristics from both. More recent works~\cite{zhang2024mimicmotion, qiu2023freenoise} have combined this strategy with a latent fusion technique. Instead of merging two fully denoised video latents, these methods integrate the predicted noise of consecutive video tiles during denoising using a position-dependent coefficient.

% Empirical observations indicate that increasing the overlap size improves the smoothness of the output video. However, if the overlap size becomes excessively large, some latent frames may encompass more than two overlapping tiles, making a linear weighting scheme designed for two tiles suboptimal. 
Naive concatenation of two independently denoised video latent representations in the temporal dimension can lead to abrupt discontinuities. A common strategy is to introduce an overlapping region where adjacent video latents are blended via a linearly weighted sum. Although increasing the overlapping size  yields smoother transitions, it also reduces temporal efficiency. Furthermore, when the overlapping region exceeds half the length of a video latent, certain frame indices may accumulate multiple independent noise predictions, potentially resulting in more inconsistencies.
To overcome these challenges and further enhance smoothness, we propose a sine weighting scheme. Unlike linear weighting, a sine function ensures smoothness in overlapping regions. This approach minimizes artifacts due to abrupt change in semantics and better preserves the consistency of the video latent, leading to a more seamless integration of the multiple denoised segments. Formally, let 
% $T$ denote the length of a latent tile in time dimension, 
$t$ denote the position of a latent frame in the final long video latent, $o$ represent the overlapping size, $n$ denote the size of a single tile. Then, the fused predicted noise $\varepsilon_t$ at video latent index $t$ is defined as:
\begin{equation}
    % \varepsilon_t = \frac{\sum_{i=j}^{k} \omega_{t - (n - o)i} \cdot \epsilon_i}{\sum \omega_{t - (n - o)i}},
    % \varepsilon_t = \frac{\sum_{i=j}^{k} \omega \cdot \epsilon_i}{\sum \omega},
    \varepsilon_t = \frac{\sum_{i=j}^{k} \bm{\omega}(t-i(n-o)) \cdot \epsilon_i}{\sum_{i=j}^{k} \bm{\omega}(t-i(n-o))},
\end{equation}
where 
\begin{equation}
    j = \lceil \frac{t-n}{n-o}\rceil,
    k = \lfloor \frac{t}{n-o}\rfloor,
\end{equation}
$\epsilon_i$ is the $i^{th}$ predicted noise tile, and
\begin{equation}
    % \omega_k = \sin(\frac{k\pi}{T} + \frac{\pi}{2T}),
    % \omega_k = \sin(\frac{k\pi}{T}) + \eta,
    \bm{\omega}(s) = \sin(\frac{s\pi}{n} + \frac{\pi}{2n}), \text{ for } s \in \mathbb{Z}_{n}
\end{equation}

% , and $\eta$ is a small positive constant that prevent the weight from being zero.

In this approach, the long noise latent $\varepsilon$ is segmented into multiple short noisy video latent tiles, each conforming to the dimensions required by the base text-to-video model. The application of weighted summation during the denoising process enables the model to achieve smooth transitions across the extended noise sequence. Furthermore, because the computation of each short noise tile during the same timestep is independent, parallel processing can be leveraged when GPU resources are abundant, thereby reducing latency.

\subsection{Long Noise Initialization}
Although latent fusion facilitates the smoothing of video latent representations between segments, it does not inherently preserve identity consistency. For instance, consider the first and last tiles of a sufficiently long video latent tensor, such that they do not share any overlapping indices. Applying latent fusion denoising to these tiles is equivalent to executing two independent text-to-video diffusion processes with distinct initial noises, which likely causes inconsistency in identity. As a result, although the final long video may exhibit smooth transitions, it does not prevent the gradual morphing of objects and scenes over time, even if the morphing process itself is smooth.

Unlike the image-to-video task, which incorporates an additional conditioning input~\cite{zhang2024mimicmotion, henschel2024streamingt2v} to constrain identity, our text-to-video approach does not rely on extra condition. However, insights can be derived from the properties of noise. As suggested in FreeNoise~\cite{qiu2023freenoise}, shuffling the noise in the temporal dimension tends to preserve object and scene identities by maintaining the low-frequency components, while primarily affecting the high-frequency details.

Inspired by FreeNoise~\cite{qiu2023freenoise}, we propose a long noise initialization method. Given a short initial noise latent with $T$ latent frames, we first repeat the initial noise $n$ times to construct an extended noise latent of length $nT$. This repetition ensures that, regardless of how the denoising tile slides across the extended long noise, each captured tile is permutation-equivalent to the others. Subsequently, for each consecutive pair of noise tiles, we shuffle the non-overlapping (stride) portion. This process introduces additional randomness into the original noise and the long noise remains a copy-and-permutation of the initial short noise, preserving most object and scene semantics. 

In addition, based on the observation that low-frequency components contribute to global features while high-frequency components contribute to local features~\cite{lu2024freelong}, we apply a Fast Fourier transform to merge the high-frequency component of the original long noise with the high-frequency component of a newly generated random noise. By progressively increasing the weight assigned to the high-frequency elements from the new random noise, we achieve a balance that preserves object and scene identity while introducing new dynamics.

% \begin{equation}
%     \epsilon_f = \textbf{IFFT}_{3D}(\textbf{FFT}_{3D}(\epsilon)\odot\mathcal{P} + w\cdot\textbf{FFT}_{3D}(\epsilon)\odot(\mathcal{1-P}) + (1-w)\cdot \textbf{FFT}_{3D}(z)\odot(\mathcal{1-P}))
% \end{equation}

\begin{algorithm}
\caption{Long Noise Initialization}
\label{alg:kstep}
\begin{algorithmic}
% \State Input:
\State $t$ := tile size
\State $o$ := overlap size
\State $n$ := number of tiles
\State $w$ := max merging factor
\newline

\State $T = \textbf{randn}(t)$
\State $T = \textbf{concat}([T] * n)$
\For {$idx$ in range(t, n*T, t - o)}
    \State $tile\_indices = \text{arange}(idx - t, idx - o)$
    \State $\textbf{shuffle}(tile\_indices)$
    \State $T[idx: idx + t - o] = T[tile\_indices]$
\EndFor
\State $\mathcal{P} = \text{low frequency filter}$
\State $z = \textbf{randn}(T)$ 
\State $T_{l} = \textbf{FFT}_\text{3D}(T) \odot \mathcal{P}$
\State $T_{h} = \textbf{FFT}_\text{3D}(T) \odot (1 - \mathcal{P})$
\State $z_{h} = \textbf{FFT}_\text{3D}(z) \odot (1 - \mathcal{P})$
\State $w = \text{linspace}(0, w, n*t)$
\State $d = \sqrt{w^2 + (1-w)^2}$
\State $T_{h} = \frac{w * T_{h} + (1-w)*z_h}{d}$
% \State $T_h = \sqrt{1-w}T_h + \sqrt{w}z_h$
\State $T_f = T_l + T_h$
\State $T = \textbf{IFFT}_\text{3D}(T_f) $
\State \Return $T$

\end{algorithmic}
\end{algorithm}

\subsection{Prompt Refining}

\begin{figure*}
\begin{center}
    \centering
    \includegraphics[width=1.0\textwidth]{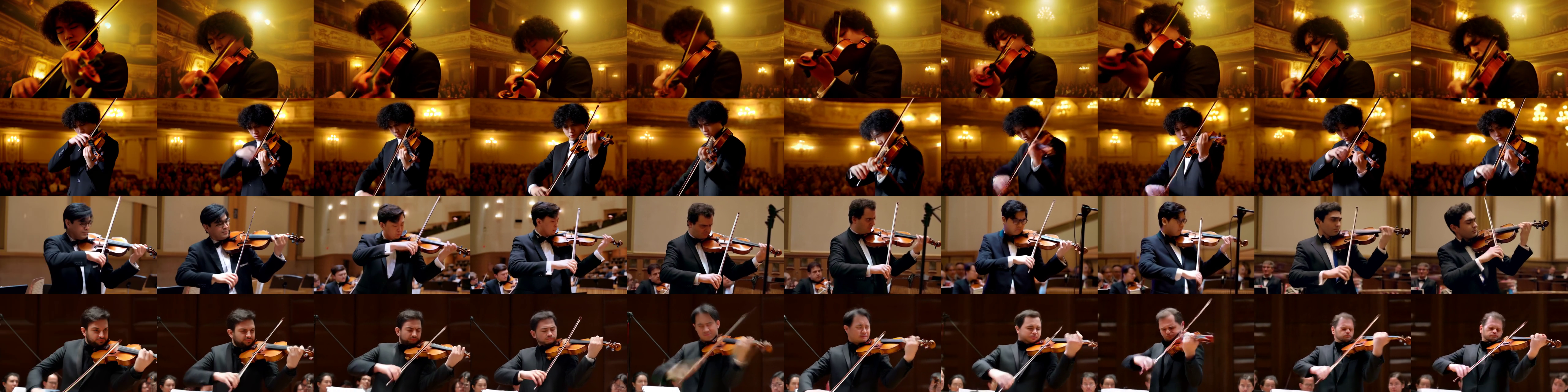}
    \caption{We demonstrated that by applying the latent fusion, long noise initialization, and prompt refining, the person identity is preserved. \textbf{Row 1 and 2}: Video examples with VideoMerge. \textbf{Row 3}: VideoMerge without latent fusion. \textbf{Row 4}: VideoMerge without prompt refining. All videos are 445-frame.}
    \label{fig:compare_violin}
\end{center}
\end{figure*}

Our observations reveal that, even when employing latent fusion and long noise initialization, preserving human facial identity becomes problematic if the input prompt is overly abstract (see Fig.~\ref{fig:compare_violin}). For example, a prompt such as ``a person is playing a violin'' is often acceptable as long as it captures the intended action; however, it does not restrict attributes such as appearance, age, or hair color. In T2V generation, such abstract prompts may induce model hallucinations. Although these hallucinations can be tolerable or even desirable in many vision generation tasks, given that it is impractical to specify every detail, the hallucination can, however, lead to obvious identity inconsistencies across latent video segments because denoising is applied independently to multiple latent tiles. Moreover, many recent large-scale video generation models~\cite{kong2024hunyuanvideo, yang2024cogvideox, yuan2024opensoraplan} incorporate large language models such as Llama~\cite{touvron2023llama} and T5~\cite{raffel2020t5} which grant better language understanding than previously commonly used CLIP~\cite{radford2021clip}, and are trained with highly detailed prompts. Thus, using overly simplistic prompts during inference may exacerbate discrepancies between training and inference conditions.

\begin{figure}
    \centering
    \includegraphics[width=1.0\linewidth]{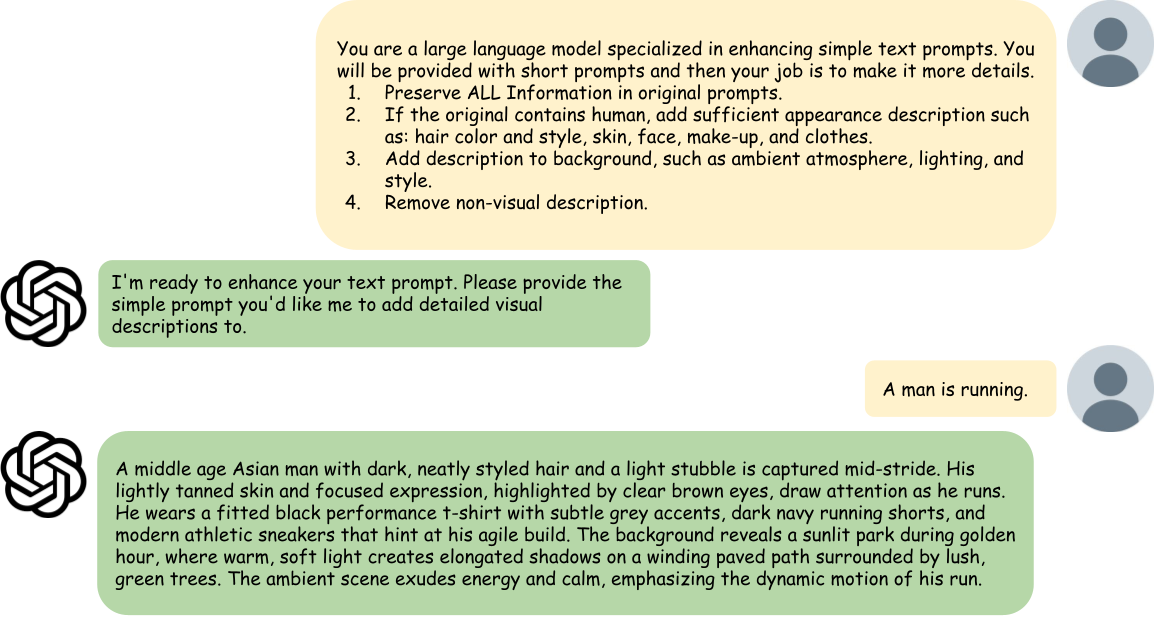}
    \caption{Prompting to a large language model to enhance short and abstract text prompts with specific requirements.}
    \label{fig:prompt_refine}
\end{figure}

\begin{figure*}
\begin{center}
    \centering
    
    \includegraphics[width=1.0\textwidth]{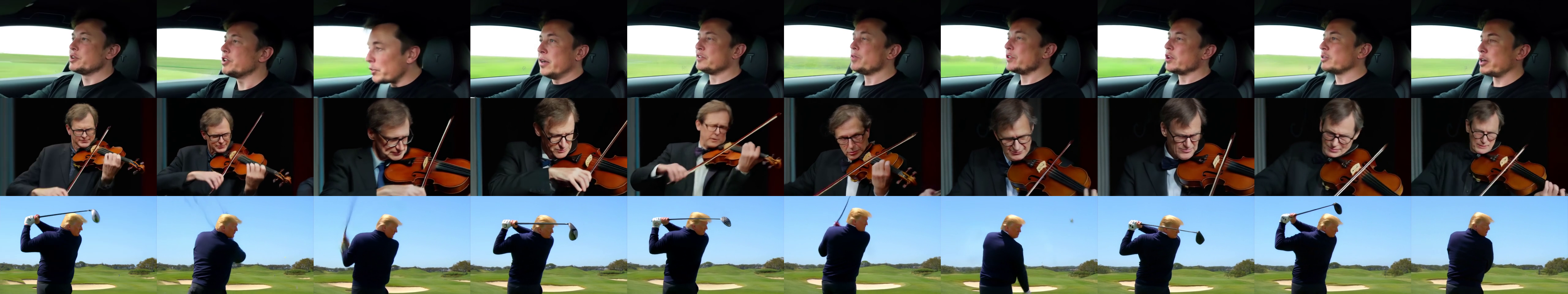}
    \footnotesize
    \caption{\textbf{Row 1 prompt}: \textit{Elon Musk is driving.} \textbf{Row 2 prompt}: \textit{Bill Gates is playing a violin.} \textbf{Row 3 prompt}: \textit{Donald Trump is playing golf.} These simple prompts with specific name token will help enhance human identity (although other details like neck tie are not preserved).}
    \label{fig:special_token}

\end{center}
\end{figure*}

We show that by providing more specific details in human appearance, the model hallucination will be restricted in a narrower range. 
We leverage large language models to refine abstract and simple prompts, enriching them with important visual details if not mentioned (e.g., hair color, age, clothing). Our experimental results indicate that these refined prompts significantly enhance the preservation of human identity in the generated videos (see Fig.~\ref{fig:compare_violin}) because more restriction in visual appearance features helps define an identity. Specifically, when generating a video containing human content, we perform prompting to a ChatGPT-O3 model~\cite{openai2024gpt4technicalreport} to create detailed and fine-grained text description (see Fig.~\ref{fig:prompt_refine}). Also, human identity can be specific if some special tokens are provided, such as names of celebrities. We also test very simple prompt with specific name, and the model is able to generate precise appearance of that indicated person (see Fig.~\ref{fig:special_token}).

\section{Experiments}
% The criteria for a “good quality” long video are not as clearly defined as those for short videos. Essential requirements might include smooth transitions, a continuous single-shot presentation (i.e., without montage), consistent human/object identity, coherent scene composition, dynamic content, abundant semantic information, and adherence to physical realism. Although these criteria are generally desirable, not all need to be simultaneously satisfied to produce a high-quality long video. For instance, drastic scene change is usually not required for landscape video. 

\begin{figure*}
\begin{center}
    \centering
    \includegraphics[width=1.0\textwidth]{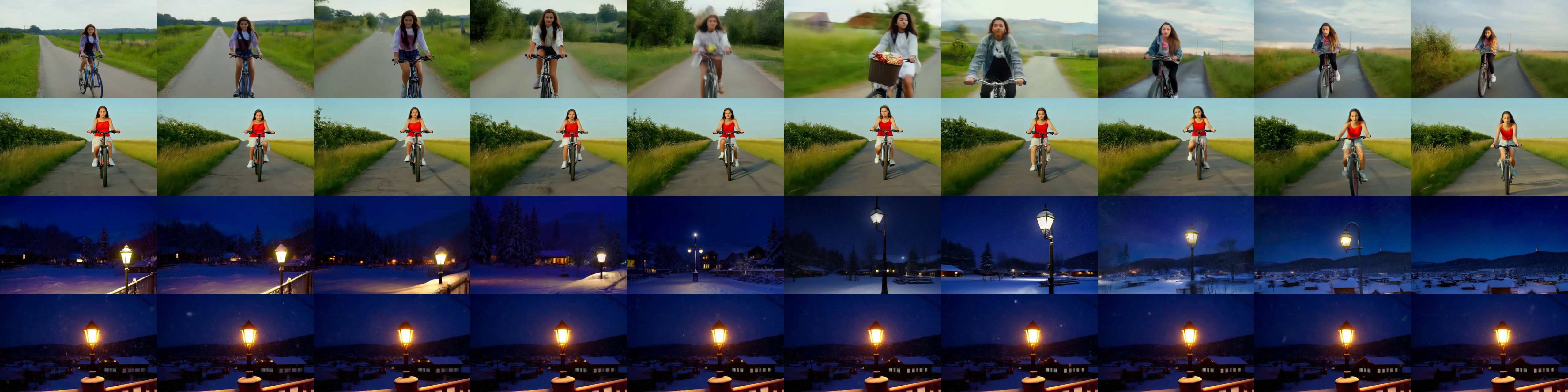}
    \caption{\textbf{Row 1 and 3}: VideoMerge without long noise initialization. These video contains artifacts such as human/object identity change. \textbf{Row 2 and 4}: Video examples with VideoMerge. All videos are 445-frame. These videos are consistent both in motion and human/object identity. }
    \label{fig:compare_noise_init}
\end{center}
\end{figure*}

We conducted experiments for long video generation based on various text prompts from VBench~\cite{huang2024vbench}, in which we divided the prompts into three categories: humans, animals, and landscapes. We compare VideoMerge with four up-to-date long video generation methods: FreeNoise~\cite{qiu2023freenoise}, FreeLong~\cite{lu2024freelong}, 
% and StreamingT2V~\cite{henschel2024streamingt2v}
FIFO~\cite{kim2024fifo}, and DiTCtrl~\cite{cai2024ditctrl}.

\begin{table*}
\small
\centering
\footnotesize
% \begin{adjustbox}{width=\textwidth}

\scalebox{1.0}{

\begin{tabular}{lccccccc}
\toprule
% {\textbf{Methods/Metrics}} &  
% \multicolumn{1}{c}{\textbf{Subject Consistency$\uparrow$}} &  \multicolumn{1}{c}{\textbf{Background Consistency$\uparrow$}}  &  \multicolumn{1}{c}{\textbf{Temporal Flickering$\uparrow$}} & \multicolumn{1}{c}{\textbf{Motion Smoothness$\uparrow$}} & \multicolumn{1}{c}{\textbf{Aesthetic Quality$\uparrow$}}  & \multicolumn{1}{c}{\textbf{Face Consistency$\uparrow$}} \\

{\textbf{Methods/Metrics}} &  
\multicolumn{1}{c}{\textbf{Sub$\uparrow$}} &  \multicolumn{1}{c}{\textbf{Back$\uparrow$}}  &  \multicolumn{1}{c}{\textbf{Temp$\uparrow$}} & \multicolumn{1}{c}{\textbf{Smooth$\uparrow$}} & \multicolumn{1}{c}{\textbf{Aes$\uparrow$}}  & \multicolumn{1}{c}{\textbf{Face$\uparrow$}} & \multicolumn{1}{c}{\textbf{FVD$\downarrow$}} \\\\

% \midrule

\cmidrule(lr){1-1} \cmidrule(lr){2-8} 

FreeNoise Human & 0.8767 & 0.9308 & 0.9271 & 0.9532 
& 0.5725 & 0.5735 & 2583\\
FreeNoise Animal  & 0.9261 & 0.9591 & 0.9392 & 0.9598 & 0.5504 & - & 2471\\
FreeNoise Landscape & 0.8960 & 0.9477 & 0.9529 & 0.9670 & 0.5818 & - & 2356\\

\cmidrule(lr){1-1} \cmidrule(lr){2-8} 
FreeLong Human & 0.9276 & 0.9549 & 0.9644 & 0.9764 
& 0.5838 & 0.7254 & 2437\\

FreeLong Animal
 & 0.9561 & 0.9621 & 0.9677 & 0.9778 
& 0.5506 & - & 1602\\

FreeLong Landscape
 & 0.9449 & 0.9658 & 0.9705 & 0.9775 
& 0.5841 & - & 1501\\

\cmidrule(lr){1-1} \cmidrule(lr){2-8} 
FIFO Human & 0.8385 & 0.9077 & 0.9076 & 0.9415 
& 0.5479 & 0.5549 & 2957
\\
FIFO Animal & 0.8811 & 0.9241 & 0.8964 & 0.9338 
& 0.5455 & - & 2171
\\
FIFO Landscape & 0.8421 & 0.9218 & 0.9535 & 0.9711 
& 0.5687 & - & 2193
\\

\cmidrule(lr){1-1} \cmidrule(lr){2-8} 
DiTCtrl Human & 0.8004 & 0.8947 & 0.9663 & 0.9813 
& 0.4871 & 0.4782 & 2939
\\
DiTCtrl Animal & 0.8307 & 0.9230 & 0.9546 & 0.9751 
& 0.5134 & - & 2796
\\
DiTCtrl Landscape & 0.8514 & 0.9167 & 0.9637 & 0.9825 
& 0.5504 & - & 2215
\\

\cmidrule(lr){1-1} \cmidrule(lr){2-8} 

VideoMerge Human (Ours) & 0.9453 & 0.9469 & 0.9832 & 0.9908 & 0.5905 & \textbf{0.9381} & 429.4\\
VideoMerge Animal (Ours)& 0.9660 & 0.9681 & 0.9863 & 0.9910 & 0.5513 & - & 450.8\\
VideoMerge Landscape (Ours)& 0.9779 & 0.9731 & 0.9919 & 0.9935 & 0.5825 & - & 357.9\\
% \midrule

\bottomrule

\end{tabular}

}

\caption{\footnotesize
    We compared our work to three up-to-date long video generation methods. \textbf{Sub} for subject consistency, \textbf{Back} for background consistency, \textbf{Temp} for temporal flickering, \textbf{Smooth} for motion smoothness, \textbf{Aes} for aesthetic quality, \textbf{Face} for face consistency, and \textbf{FVD} for Frechet Video Distance.
}
\label{tab:vbench_metric}

\end{table*}

\subsection{Metrics and Results}
To evaluate the quality of the generated videos, we selected five dimensions from VBench's~\cite{huang2024vbench} Video Quality category, including Subject Consistency, Background Consistency, Temporal Flickering, Motion Smoothness, and Aesthetic Quality. Specifically, \textbf{Subject Consistency} is defined as the average cosine similarity between the DINO~\cite{caron2021dino} features of every pair of consecutive frames; \textbf{Background Consistency} is similar to Subject Consistency except that the DINO features are replaced with CLIP~\cite{radford2021clip} features; \textbf{Temporal Flickering} is defined as the mean absolute difference between frames in pixel space. As for \textbf{Motion Smoothness}, a pretrained video frame interpolation model AMT~\cite{li2023amt} is first utilized to reconstruct all odd-number frames from the even-number frames, after which a Mean Absolute Error is calculated between the ground-truth and reconstructed odd-number frames.
\textbf{Aesthetic Quality}, reflecting frame-level aesthetic aspects, is calculated with the LAION aesthetic predictor~\cite{LAIONaes} in all frames.

In addition, as we also pay attention to the preservation of human identity, we take advantage of Face Recognition~\cite{ageitgey2018facerecog} to evaluate the consistency of human faces among video frames. For a human video, we first choose its first frame and encode the human face information from that frame, and then for later frames, we encode each of them and compare with the first frame encoding. The \textbf{Face Consistency} is defined as:
\begin{equation}
    \mathcal{C}_v = \frac{\sum_{f\in v[1:]}||\mathbf{R}(v_0) - \mathbf{R}(f))|| < \tau}{||v|| - 1}
\end{equation}
where $\mathbf{R}$ is the face recognition model, $v$ is a video, $v_0$ is the first frame of $v$, and $\tau$ is the tolerance constant.

In Table.~\ref{tab:vbench_metric}, we show quantitative results of VideoMerge against other state-of-the-art long video generation methods. We demonstrated that our proposed method achieves superior performance to other methods, especially for the face consistency metrics. We also show qualitative results of our method and other methods in the human, animal and landscape categories (see Fig.~\ref{fig:qualitative_results_dancer}, ~\ref{fig:qualitative_results_tiger}, ~\ref{fig:qualitative_results_waterfall}).
% Fig~\ref{fig:qualitative_results_dancer}

We further measure Frechet Video Distance(FVD)~\cite{unterthiner2019fvd} that describe similarity between two video sets. For each method and category, we generate a short video set with same prompts and random seed, as a reference set to the long video set.

\subsection{Implementation Details}
We adopt Hunyuan~\cite{kong2024hunyuanvideo} as our base model, and set the single tile noise tensor shape to [1, 16, 16, 40, 64], which corresponds to a latent representation of a $61\times320\times512$ video in pixel space. We set the overlapping size to 12, which grants a smooth transition between adjacent latent video tiles. As the overlapping size is greater than half of the tile length, some frame indices consists of more than two denoising tiles. Specifically, it takes 22 minutes to generate a 445-frame video with 320p resolution, with 45GB GPU memory usage. 

\subsection{Ablation Studies}

To evaluate the effectiveness of each component in our proposed VideoMerge framework, we conducted a series of ablation experiments. In these studies, we selectively disable each of latent fusion, prompt refining, and long noise initialization and compare the results to those produced by the fully integrated method. This analysis allows us to quantify the contribution of each component toward maintaining both semantic and identity consistency across generated videos.

Fig.~\ref{fig:compare_violin} presents an example of 445-frame videos that depict a person playing the violin. In the first two rows, we provide results from the fully exploited VideoMerge method using different random seeds. These examples robustly preserve both the human subject’s identity and the overall scene consistency.
% , thereby demonstrating the reproducibility and stability of our approach under varying initialization conditions.

\begin{figure}
    \centering
    \includegraphics[width=1.0\linewidth]{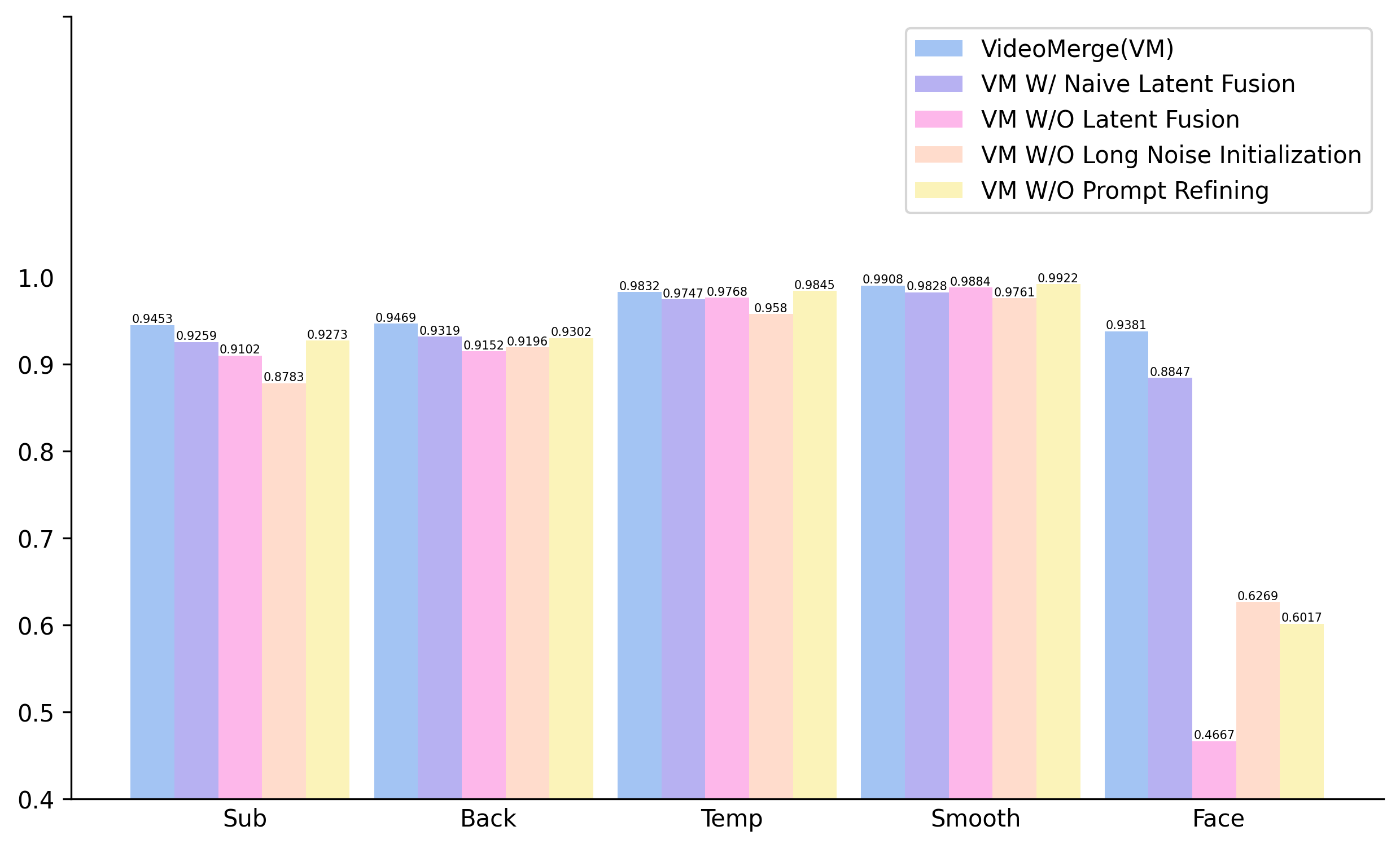}
    \caption{We ablate each component of VideoMerge. There are drops in performance especially for face consistency. }
    \label{fig:ablation}
\end{figure}

% \begin{table}
% \small
% \centering
% \footnotesize
% % \begin{adjustbox}{width=\textwidth}

% \scalebox{1.0}{

% \begin{tabular}{lccccc}
% \toprule

% {\textbf{Methods/Metrics}} &  
% \multicolumn{1}{c}{\textbf{Sub$\uparrow$}} &  \multicolumn{1}{c}{\textbf{Back$\uparrow$}}  &  \multicolumn{1}{c}{\textbf{Temp$\uparrow$}} & \multicolumn{1}{c}{\textbf{Smooth$\uparrow$}} & 
% % \multicolumn{1}{c}{\textbf{Aes$\uparrow$}}  & 
% \multicolumn{1}{c}{\textbf{Face$\uparrow$}} \\

% % \midrule

% \cmidrule(lr){1-1} \cmidrule(lr){2-6} 

% VideoMerge & 0.9453 & 0.9469 & 0.9832 & 0.9908 & \textbf{0.9381} \\
% W/O Fusion & 0.9102 & 0.9152 & 0.9768 & 0.9884 & 0.4667\\
% W/O Init & 0.8783 & 0.9196 & 0.9580 & 0.9761 & 0.6269\\
% W/O Prompt & 0.9273 & 0.9302 & 0.9845 & 0.9922 & 0.6017\\

% \bottomrule

% \end{tabular}

% }

% \caption{\footnotesize
%     We ablate each component of VideoMerge. There is a drop in performance especially for face consistency. W/O Fusion: without Latent Fusion; W/O Init: without long noise initialization; W/O Prompt: without prompt refining.
% }
% \label{tab:ablation}

% \end{table}

When the latent fusion is disabled (as shown in Fig.~\ref{fig:compare_violin} row 3), significant degradations become apparent. Not only does the subject’s identity undergo undesirable fluctuations, transitioning from a man with glasses to a man with curly hair and then reverting to the original appearance, the background also experiences abrupt and incongruous changes, such as the sudden appearance of a microphone and an unexpected audience seat on the second floor. These observations underscore the pivotal role of latent fusion in ensuring smooth transitions and coherent blending of adjacent denoising tiles.

Similarly, the ablation of the prompt refining stage (illustrated in Fig.~\ref{fig:compare_violin} row 4) reveals that while the background and overall human body structure are reasonably preserved, high-level facial details suffer from instability. The human face exhibits inconsistent attributes, fluctuating between different states (e.g., a man with a brown beard, then a man with no beard, and subsequently reverting back to a brown beard). This inconsistency highlights the importance of the prompt refining mechanism in enforcing fine-grained semantic control over high-level visual features.

Further experiments, as depicted in Fig.~\ref{fig:compare_noise_init}, demonstrate that the long noise initialization strategy is crucial for maintaining object identity within dynamic scenes. Without this component, elements such as the identity of the girl and the shape of the lamp are not consistently preserved throughout the video. This indicates that the long noise initialization not only contributes to dynamic variability but also plays an essential role in the reliable preservation of detailed visual cues.

We also provide quantitative results for ablations of each component in our methods (see Fig.~\ref{fig:ablation}), demonstrating their effectiveness. The latent fusion module ensures semantic continuity across video segments, the prompt refining stage enhances high-level detail consistency, and the long noise initialization strategy guarantees the faithful preservation of both dynamic and static scene elements. The integration of these components in VideoMerge is thus effective in achieving superior long-video synthesis performance while maintaining both visual coherence and identity fidelity.

% \subsubsection*{Special Token}

% In our previous experiments, we have demonstrated that providing fine-grain description on human appearance will help enhance identity consistency. Specifically, we also tried to provide abstract prompt but with very specific token, such as names of celebrities, and found that such special token can force the model to generate the exact appearance of that person (see Fig.~\ref{fig:special_token}). 
% Such observation provides a possibility to apply methods like Dreambooth~\cite{ruiz2023dreambooth} or Textual Inversion~\cite{gal2022textiversion} for identity specific finetuning.
\section{Conclusion}
We present VideoMerge, a training‐free method that effectively extends the capabilities of pretrained text‐to‐video diffusion models for consistent long video generation with limited GPU resources. 
% Our method addresses both feasibility and quality challenges by leveraging a sliding window approach that conforms to the original training configuration while mitigating GPU memory constraints. 
Specifically, our contributions include a novel latent fusion technique that ensures smooth transitions between video segments, a long noise initialization strategy that harmonizes consistency with dynamic variability, and a prompt refining mechanism that robustly preserves human identity across frames.

VideoMerge not only maintains the high expressiveness and quality inherent to the pretrained models but also significantly enhances temporal coherence and detail preservation in extended video sequences. By capitalizing on the strengths of existing diffusion models without the need for additional training, our approach offers a computationally efficient solution to long consistent video generation task. 
% Our work lays a foundation for future research in long consistent video generation, opening avenues for further refinement and adaptation of diffusion-based techniques to broader applications in the computer vision domain.
{
    \small
    \bibliographystyle{ieeenat_fullname}
    \bibliography{main}

\begin{thebibliography}{35}
\providecommand{\natexlab}[1]{#1}
\providecommand{\url}[1]{\texttt{#1}}
\expandafter\ifx\csname urlstyle\endcsname\relax
  \providecommand{\doi}[1]{doi: #1}\else
  \providecommand{\doi}{doi: \begingroup \urlstyle{rm}\Url}\fi

\bibitem[Bar-Tal et~al.(2024)Bar-Tal, Chefer, Tov, Herrmann, Paiss, Zada, Ephrat, Hur, Liu, Raj, et~al.]{bar2024lumiere}
Omer Bar-Tal, Hila Chefer, Omer Tov, Charles Herrmann, Roni Paiss, Shiran Zada, Ariel Ephrat, Junhwa Hur, Guanghui Liu, Amit Raj, et~al.
\newblock Lumiere: A space-time diffusion model for video generation.
\newblock In \emph{SIGGRAPH Asia 2024 Conference Papers}, pages 1--11, 2024.

\bibitem[Blattmann et~al.(2023)Blattmann, Dockhorn, Kulal, Mendelevitch, Kilian, Lorenz, Levi, English, Voleti, Letts, et~al.]{blattmann2023stablevideodiffusion}
Andreas Blattmann, Tim Dockhorn, Sumith Kulal, Daniel Mendelevitch, Maciej Kilian, Dominik Lorenz, Yam Levi, Zion English, Vikram Voleti, Adam Letts, et~al.
\newblock Stable video diffusion: Scaling latent video diffusion models to large datasets.
\newblock \emph{arXiv preprint arXiv:2311.15127}, 2023.

\bibitem[Butterworth et~al.(1930)]{butterworth1930theory}
Stephen Butterworth et~al.
\newblock On the theory of filter amplifiers.
\newblock \emph{Wireless Engineer}, 7\penalty0 (6):\penalty0 536--541, 1930.

\bibitem[Cai et~al.(2024)Cai, Cun, Li, Liu, Zhang, Zhang, Shan, and Yue]{cai2024ditctrl}
Minghong Cai, Xiaodong Cun, Xiaoyu Li, Wenze Liu, Zhaoyang Zhang, Yong Zhang, Ying Shan, and Xiangyu Yue.
\newblock Ditctrl: Exploring attention control in multi-modal diffusion transformer for tuning-free multi-prompt longer video generation.
\newblock \emph{arXiv preprint arXiv:2412.18597}, 2024.

\bibitem[Caron et~al.(2021)Caron, Touvron, Misra, J{\'e}gou, Mairal, Bojanowski, and Joulin]{caron2021dino}
Mathilde Caron, Hugo Touvron, Ishan Misra, Herv{\'e} J{\'e}gou, Julien Mairal, Piotr Bojanowski, and Armand Joulin.
\newblock Emerging properties in self-supervised vision transformers.
\newblock In \emph{Proceedings of the IEEE/CVF international conference on computer vision}, pages 9650--9660, 2021.

\bibitem[Chen et~al.(2024)Chen, Zhang, Cun, Xia, Wang, Weng, and Shan]{chen2024videocrafter2}
Haoxin Chen, Yong Zhang, Xiaodong Cun, Menghan Xia, Xintao Wang, Chao Weng, and Ying Shan.
\newblock Videocrafter2: Overcoming data limitations for high-quality video diffusion models.
\newblock In \emph{Proceedings of the IEEE/CVF Conference on Computer Vision and Pattern Recognition}, pages 7310--7320, 2024.

\bibitem[Esser et~al.(2024)Esser, Kulal, Blattmann, Entezari, M{\"u}ller, Saini, Levi, Lorenz, Sauer, Boesel, et~al.]{esser2024sd3flow}
Patrick Esser, Sumith Kulal, Andreas Blattmann, Rahim Entezari, Jonas M{\"u}ller, Harry Saini, Yam Levi, Dominik Lorenz, Axel Sauer, Frederic Boesel, et~al.
\newblock Scaling rectified flow transformers for high-resolution image synthesis.
\newblock In \emph{Forty-first international conference on machine learning}, 2024.

\bibitem[Geitgey(2018)]{ageitgey2018facerecog}
Adam Geitgey.
\newblock Face recognition.
\newblock \url{https://github.com/ageitgey/face_recognition}, 2018.
\newblock Accessed: 2025-02-27.

\bibitem[Goodfellow et~al.(2014)Goodfellow, Pouget-Abadie, Mirza, Xu, Warde-Farley, Ozair, Courville, and Bengio]{goodfellow2014generative}
Ian Goodfellow, Jean Pouget-Abadie, Mehdi Mirza, Bing Xu, David Warde-Farley, Sherjil Ozair, Aaron Courville, and Yoshua Bengio.
\newblock Generative adversarial nets.
\newblock \emph{Advances in neural information processing systems}, 27, 2014.

\bibitem[Guo et~al.(2023)Guo, Yang, Rao, Liang, Wang, Qiao, Agrawala, Lin, and Dai]{guo2023animatediff}
Yuwei Guo, Ceyuan Yang, Anyi Rao, Zhengyang Liang, Yaohui Wang, Yu Qiao, Maneesh Agrawala, Dahua Lin, and Bo Dai.
\newblock Animatediff: Animate your personalized text-to-image diffusion models without specific tuning.
\newblock \emph{arXiv preprint arXiv:2307.04725}, 2023.

\bibitem[Henschel et~al.(2024)Henschel, Khachatryan, Hayrapetyan, Poghosyan, Tadevosyan, Wang, Navasardyan, and Shi]{henschel2024streamingt2v}
Roberto Henschel, Levon Khachatryan, Daniil Hayrapetyan, Hayk Poghosyan, Vahram Tadevosyan, Zhangyang Wang, Shant Navasardyan, and Humphrey Shi.
\newblock Streamingt2v: Consistent, dynamic, and extendable long video generation from text.
\newblock \emph{arXiv preprint arXiv:2403.14773}, 2024.

\bibitem[Ho et~al.(2020)Ho, Jain, and Abbeel]{ho2020denoising}
Jonathan Ho, Ajay Jain, and Pieter Abbeel.
\newblock Denoising diffusion probabilistic models.
\newblock \emph{Advances in neural information processing systems}, 33:\penalty0 6840--6851, 2020.

\bibitem[Ho et~al.(2022)Ho, Salimans, Gritsenko, Chan, Norouzi, and Fleet]{ho2022video}
Jonathan Ho, Tim Salimans, Alexey Gritsenko, William Chan, Mohammad Norouzi, and David~J Fleet.
\newblock Video diffusion models.
\newblock \emph{Advances in Neural Information Processing Systems}, 35:\penalty0 8633--8646, 2022.

\bibitem[Huang et~al.(2024)Huang, He, Yu, Zhang, Si, Jiang, Zhang, Wu, Jin, Chanpaisit, et~al.]{huang2024vbench}
Ziqi Huang, Yinan He, Jiashuo Yu, Fan Zhang, Chenyang Si, Yuming Jiang, Yuanhan Zhang, Tianxing Wu, Qingyang Jin, Nattapol Chanpaisit, et~al.
\newblock Vbench: Comprehensive benchmark suite for video generative models.
\newblock In \emph{Proceedings of the IEEE/CVF Conference on Computer Vision and Pattern Recognition}, pages 21807--21818, 2024.

\bibitem[Kim et~al.(2024)Kim, Kang, Choi, and Han]{kim2024fifo}
Jihwan Kim, Junoh Kang, Jinyoung Choi, and Bohyung Han.
\newblock Fifo-diffusion: Generating infinite videos from text without training.
\newblock \emph{arXiv preprint arXiv:2405.11473}, 2024.

\bibitem[Kong et~al.(2024)Kong, Tian, Zhang, Min, Dai, Zhou, Xiong, Li, Wu, Zhang, Wu, Lin, Wang, Wang, Li, Huang, Yang, Tan, Wang, Song, Bai, Wu, Xue, Wang, Yuan, Wang, Liu, Li, Li, Wang, Yu, Deng, Li, Long, Chen, Cui, Peng, Yu, He, Xu, Zhou, Xu, Tao, Lu, Liu, Zhou, Wang, Yang, Wang, Liu, , Jiang, and along~with Caesar~Zhong]{kong2024hunyuanvideo}
Weijie Kong, Qi Tian, Zijian Zhang, Rox Min, Zuozhuo Dai, Jin Zhou, Jiangfeng Xiong, Xin Li, Bo Wu, Jianwei Zhang, Kathrina Wu, Qin Lin, Aladdin Wang, Andong Wang, Changlin Li, Duojun Huang, Fang Yang, Hao Tan, Hongmei Wang, Jacob Song, Jiawang Bai, Jianbing Wu, Jinbao Xue, Joey Wang, Junkun Yuan, Kai Wang, Mengyang Liu, Pengyu Li, Shuai Li, Weiyan Wang, Wenqing Yu, Xinchi Deng, Yang Li, Yanxin Long, Yi Chen, Yutao Cui, Yuanbo Peng, Zhentao Yu, Zhiyu He, Zhiyong Xu, Zixiang Zhou, Zunnan Xu, Yangyu Tao, Qinglin Lu, Songtao Liu, Dax Zhou, Hongfa Wang, Yong Yang, Di Wang, Yuhong Liu, , Jie Jiang, and along~with Caesar~Zhong.
\newblock Hunyuanvideo: A systematic framework for large video generative models, 2024.

\bibitem[Lab and etc.(2024)]{yuan2024opensoraplan}
PKU-Yuan Lab and Tuzhan~AI etc.
\newblock Open-sora-plan, 2024.

\bibitem[LAION-AI(2022)]{LAIONaes}
LAION-AI.
\newblock aesthetic-predictor.
\newblock \url{https://github.com/LAION-AI/aesthetic-predictor}, 2022.

\bibitem[Li et~al.(2023)Li, Zhu, Han, Hou, Guo, and Cheng]{li2023amt}
Zhen Li, Zuo-Liang Zhu, Ling-Hao Han, Qibin Hou, Chun-Le Guo, and Ming-Ming Cheng.
\newblock Amt: All-pairs multi-field transforms for efficient frame interpolation.
\newblock In \emph{Proceedings of the IEEE/CVF Conference on Computer Vision and Pattern Recognition}, pages 9801--9810, 2023.

\bibitem[Lu et~al.(2024)Lu, Liang, Zhu, and Yang]{lu2024freelong}
Yu Lu, Yuanzhi Liang, Linchao Zhu, and Yi Yang.
\newblock Freelong: Training-free long video generation with spectralblend temporal attention.
\newblock \emph{arXiv preprint arXiv:2407.19918}, 2024.

\bibitem[OpenAI et~al.(2024)OpenAI, Achiam, Adler, et~al.]{openai2024gpt4technicalreport}
OpenAI, Josh Achiam, Steven Adler, et~al.
\newblock Gpt-4 technical report, 2024.

\bibitem[Peebles and Xie(2023)]{peebles2023dit}
William Peebles and Saining Xie.
\newblock Scalable diffusion models with transformers.
\newblock In \emph{Proceedings of the IEEE/CVF International Conference on Computer Vision}, pages 4195--4205, 2023.

\bibitem[Qiu et~al.(2023)Qiu, Xia, Zhang, He, Wang, Shan, and Liu]{qiu2023freenoise}
Haonan Qiu, Menghan Xia, Yong Zhang, Yingqing He, Xintao Wang, Ying Shan, and Ziwei Liu.
\newblock Freenoise: Tuning-free longer video diffusion via noise rescheduling.
\newblock \emph{arXiv preprint arXiv:2310.15169}, 2023.

\bibitem[Radford et~al.(2021)Radford, Kim, Hallacy, Ramesh, Goh, Agarwal, Sastry, Askell, Mishkin, Clark, et~al.]{radford2021clip}
Alec Radford, Jong~Wook Kim, Chris Hallacy, Aditya Ramesh, Gabriel Goh, Sandhini Agarwal, Girish Sastry, Amanda Askell, Pamela Mishkin, Jack Clark, et~al.
\newblock Learning transferable visual models from natural language supervision.
\newblock In \emph{International conference on machine learning}, pages 8748--8763. PMLR, 2021.

\bibitem[Raffel et~al.(2020)Raffel, Shazeer, Roberts, Lee, Narang, Matena, Zhou, Li, and Liu]{raffel2020t5}
Colin Raffel, Noam Shazeer, Adam Roberts, Katherine Lee, Sharan Narang, Michael Matena, Yanqi Zhou, Wei Li, and Peter~J Liu.
\newblock Exploring the limits of transfer learning with a unified text-to-text transformer.
\newblock \emph{Journal of machine learning research}, 21\penalty0 (140):\penalty0 1--67, 2020.

\bibitem[Ronneberger et~al.(2015)Ronneberger, Fischer, and Brox]{ronneberger2015unet}
Olaf Ronneberger, Philipp Fischer, and Thomas Brox.
\newblock U-net: Convolutional networks for biomedical image segmentation.
\newblock In \emph{Medical image computing and computer-assisted intervention--MICCAI 2015: 18th international conference, Munich, Germany, October 5-9, 2015, proceedings, part III 18}, pages 234--241. Springer, 2015.

\bibitem[Touvron et~al.(2023)Touvron, Lavril, Izacard, Martinet, Lachaux, Lacroix, Rozi{\`e}re, Goyal, Hambro, Azhar, et~al.]{touvron2023llama}
Hugo Touvron, Thibaut Lavril, Gautier Izacard, Xavier Martinet, Marie-Anne Lachaux, Timoth{\'e}e Lacroix, Baptiste Rozi{\`e}re, Naman Goyal, Eric Hambro, Faisal Azhar, et~al.
\newblock Llama: Open and efficient foundation language models.
\newblock \emph{arXiv preprint arXiv:2302.13971}, 2023.

\bibitem[Unterthiner et~al.(2019)Unterthiner, Van~Steenkiste, Kurach, Marinier, Michalski, and Gelly]{unterthiner2019fvd}
Thomas Unterthiner, Sjoerd Van~Steenkiste, Karol Kurach, Rapha{\"e}l Marinier, Marcin Michalski, and Sylvain Gelly.
\newblock Fvd: A new metric for video generation.
\newblock 2019.

\bibitem[Wang et~al.(2023)Wang, Chen, Song, Ye, Liu, and Li]{wang2023genlvideo}
Fu-Yun Wang, Wenshuo Chen, Guanglu Song, Han-Jia Ye, Yu Liu, and Hongsheng Li.
\newblock Gen-l-video: Multi-text to long video generation via temporal co-denoising.
\newblock \emph{arXiv preprint arXiv:2305.18264}, 2023.

\bibitem[Wang et~al.(2024)Wang, Chen, Ma, Zhou, Huang, Wang, Yang, He, Yu, Yang, et~al.]{wang2024lavie}
Yaohui Wang, Xinyuan Chen, Xin Ma, Shangchen Zhou, Ziqi Huang, Yi Wang, Ceyuan Yang, Yinan He, Jiashuo Yu, Peiqing Yang, et~al.
\newblock Lavie: High-quality video generation with cascaded latent diffusion models.
\newblock \emph{International Journal of Computer Vision}, pages 1--20, 2024.

\bibitem[Yang et~al.(2024)Yang, Teng, Zheng, Ding, Huang, Xu, Yang, Hong, Zhang, Feng, et~al.]{yang2024cogvideox}
Zhuoyi Yang, Jiayan Teng, Wendi Zheng, Ming Ding, Shiyu Huang, Jiazheng Xu, Yuanming Yang, Wenyi Hong, Xiaohan Zhang, Guanyu Feng, et~al.
\newblock Cogvideox: Text-to-video diffusion models with an expert transformer.
\newblock \emph{arXiv preprint arXiv:2408.06072}, 2024.

\bibitem[Yu et~al.(2023)Yu, Lezama, Gundavarapu, Versari, Sohn, Minnen, Cheng, Birodkar, Gupta, Gu, et~al.]{yu2023magvit2}
Lijun Yu, Jos{\'e} Lezama, Nitesh~B Gundavarapu, Luca Versari, Kihyuk Sohn, David Minnen, Yong Cheng, Vighnesh Birodkar, Agrim Gupta, Xiuye Gu, et~al.
\newblock Language model beats diffusion--tokenizer is key to visual generation.
\newblock \emph{arXiv preprint arXiv:2310.05737}, 2023.

\bibitem[Zhang et~al.(2024)Zhang, Gu, Wang, Wang, Cheng, Zhu, and Zou]{zhang2024mimicmotion}
Yuang Zhang, Jiaxi Gu, Li-Wen Wang, Han Wang, Junqi Cheng, Yuefeng Zhu, and Fangyuan Zou.
\newblock Mimicmotion: High-quality human motion video generation with confidence-aware pose guidance.
\newblock \emph{arXiv preprint arXiv:2406.19680}, 2024.

\bibitem[Zheng et~al.(2024)Zheng, Peng, Yang, Shen, Li, Liu, Zhou, Li, and You]{opensora}
Zangwei Zheng, Xiangyu Peng, Tianji Yang, Chenhui Shen, Shenggui Li, Hongxin Liu, Yukun Zhou, Tianyi Li, and Yang You.
\newblock Open-sora: Democratizing efficient video production for all, 2024.

\bibitem[Zhou et~al.(2024)Zhou, Wang, Cai, and Yang]{allegro2024}
Yuan Zhou, Qiuyue Wang, Yuxuan Cai, and Huan Yang.
\newblock Allegro: Open the black box of commercial-level video generation model.
\newblock \emph{arXiv preprint arXiv:2410.15458}, 2024.

\end{thebibliography}
}

\twocolumn[{
\renewcommand\twocolumn[1][]{#1}%
\maketitlesupplementary

\begin{center}
    \vspace{-20pt}
    \includegraphics[
    % width=0.58\textwidth,
    % height=0.5\textwidth
    width=1.0\textwidth,
    height=0.3\textwidth
    ]{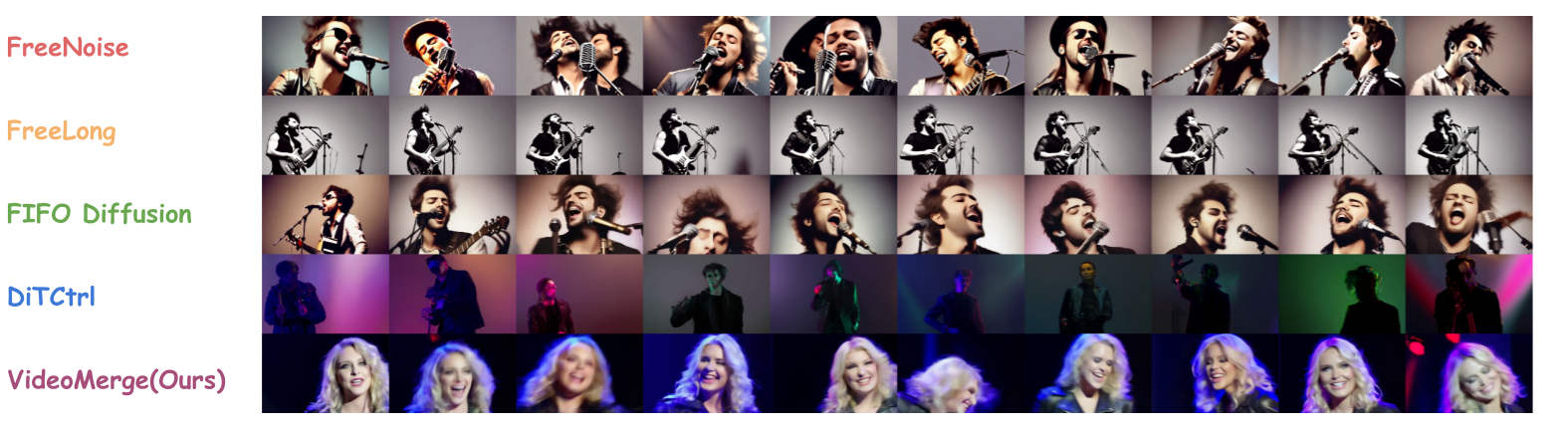}
    \captionof{figure}{Original Prompt: \textit{A singer on the stage.}}
    
    \includegraphics[
    % width=0.58\textwidth,
    % height=0.5\textwidth
    width=1.0\textwidth,
    height=0.3\textwidth
    ]{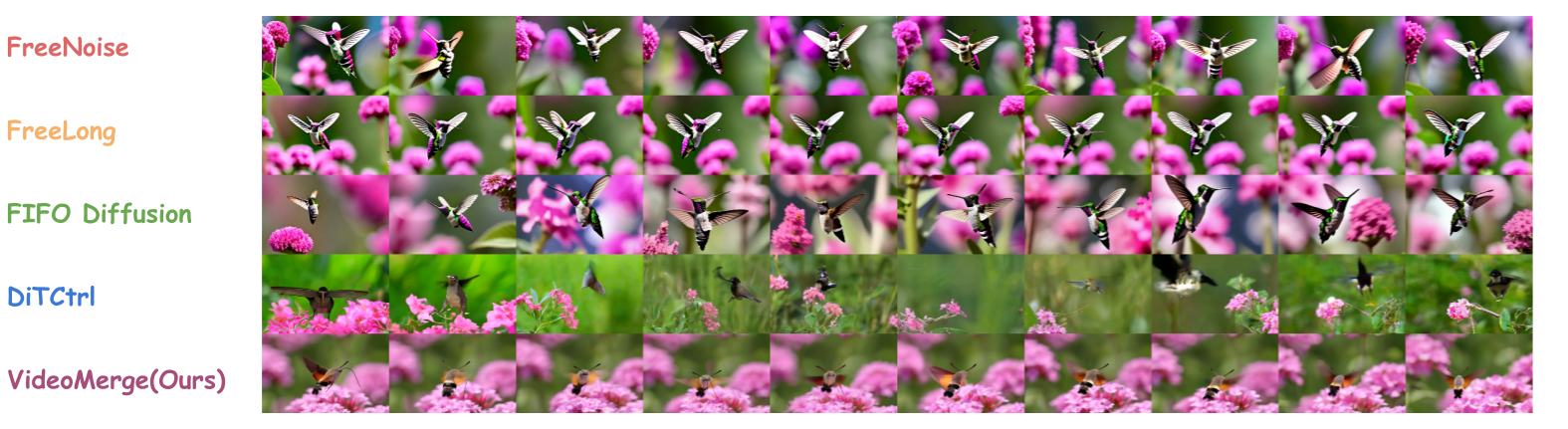}
    \captionof{figure}{Original Prompt: \textit{Hummingbird hawk moth flying near pink flowers.}}

    % \vspace{-20pt}
    \includegraphics[
    % width=0.58\textwidth,
    % height=0.5\textwidth
    width=1.0\textwidth,
    height=0.3\textwidth
    ]{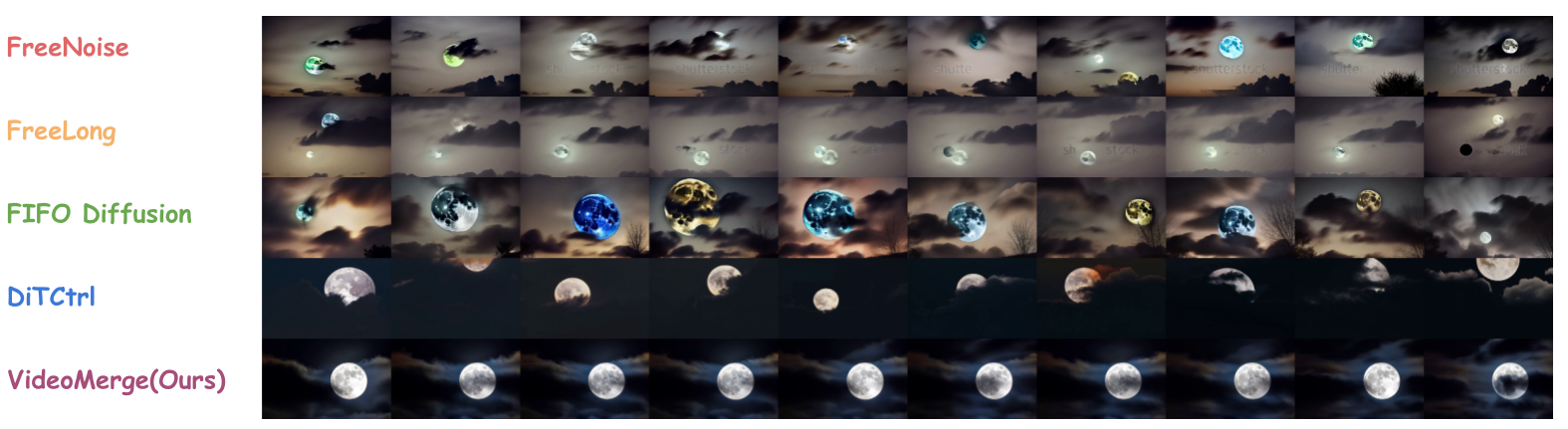}
    
\vspace{-0.4cm}
\captionof{figure}{Comparison between our proposed \textit{VideoMerge} and other state-of-the-art methods. Original prompt: \textit{``Dark clouds over shadowing the full moon.''} Our method is able to preserve consistency in human identity in terms of face, clothes, hair style. \textbf{More video} examples are in the supplementary files.
}
    \vspace{3pt}
    \label{fig:singer}
    \end{center}%
}]

\section*{A. More Qualitative Results in Long Video}

With our VideoMerge method, we successfully extend the capability of base Hunyuan Text-to-video model~\cite{kong2024hunyuanvideo} to generate long videos. For a detailed visualization, please refer to the local \textit{\href{./gallery.html}{gallery.html}} file included in the supplementary folder, where we provide examples of high quality long videos in human, animal and landscape categories. We also present some high resolution ($1280\times 720$) videos, please zoom in for better view.

\section*{B. More Implementation Details}
For inference, we use FlowMatchEulerDiscreteScheduler~\cite{esser2024sd3flow} on HunyuanDiT~\cite{kong2024hunyuanvideo} model, with 30 denoising steps. The resolution corresponding to each denoising tile is $61\times320\times512$, which corresponds to a $1\times16\times16\times40\times64$ tensor in latent space. We set overlapping size to 12, and thus for a 445-frame(7 denoising tiles) video, we need to perform denoising on 25 individual noise tiles. Our method require 45GB GPU memory to generate a 445-frame $320\times512$ resolution video in around 22min. Specifically, given a single denoising step 1.78s on a single NVIDIA H100 GPU, and total denoising steps 750. Multiple GPU parallelism can help accelerate generation because each latent tile is denoised independently within the same timestep.

In quantitative results, for each categories in human, animal and landscape, and for each method, we generate long videos with prompts from VBench~\cite{huang2024vbench}. VideoMerge (ours), FreeNoise~\cite{qiu2023freenoise}, FIFO~\cite{kim2024fifo}, and DitCtrl~\cite{cai2024ditctrl} videos are all truncated into 445-frame, while FreeLong~\cite{lu2024freelong} videos are 256-frame due to memory restriction(the extended attention mask for FreeLong will cause out of memory problem when the number of frames is too large). 

For frequency decomposition filter, we choose the Butterworth Filter~\cite{butterworth1930theory}, and set the threshold for spatial and temporal frequency to default 0.25, and the order of filter to 4.

We merge high frequency of a new long noise tensor into the original long noise with progressively increasing weights from 0 to 0.1. 

We directly apply metrics in VBench~\cite{huang2024vbench} on these long videos and collect results showing in the experiments session. 
For Frechet Video Distance (FVD)~\cite{unterthiner2019fvd}, we use a short video set generated by each method with same prompt and same random seed, as a reference set of that method. To ensure fairness, we slice our generated long videos in to short ones, and then compare with the reference set. Furthermore, as each base model has distinct default video tile length, we also truncate long videos so that the long-to-short length ratio are the same as a minimum of number of tiles. To be specific, for VideoMerge, we split 445-frames video into \textbf{7} short videos, each with length 61 frames because the base model has default length 61; for FreeNoise\cite{qiu2023freenoise}, FreeLong~\cite{lu2024freelong} and FIFO~\cite{kim2024fifo}, we only choose the first 112 frames and slice into \textbf{7} 16-frame short videos because the base model has default length 16; and for DiTCtrl~\cite{cai2024ditctrl}, \textbf{7} 49-frame videos. In such setting, the FVD is able to describe in a fair way similarity with the long video set and the short video set generated with the same base method.
\end{document}